# Sequential Interpretability: Methods, Applications, and Future Direction for Understanding Deep Learning Models in the Context of Sequential Data

Benjamin Shickel[1,3], Parisa Rashidi[2,3]

[1] Department of Computer and Information Science and Engineering
[2] Department of Biomedical Engineering
[3] University of Florida, Gainesville, Florida, United States

## ABSTRACT

Deep learning continues to revolutionize an ever-growing number of critical application areas including healthcare, transportation, finance, and basic sciences. Despite their increased predictive power, model transparency and human explainability remain a significant challenge due to the "black box" nature of modern deep learning models. In many cases the desired balance between interpretability and performance is predominately task specific. Human-centric domains such as healthcare necessitate a renewed focus on understanding how and why these frameworks are arriving at critical and potentially life-or-death decisions. Given the quantity of research and empirical successes of deep learning for computer vision, most of the existing interpretability research has focused on image processing techniques. Comparatively, less attention has been paid to interpreting deep learning frameworks using sequential data. Given recent deep learning advancements in highly sequential domains such as natural language processing and physiological signal processing, the need for deep sequential explanations is at an all-time high. In this paper, we review current techniques for interpreting deep learning techniques involving sequential data, identify similarities to non-sequential methods, and discuss current limitations and future avenues of sequential interpretability research.

## I. INTRODUCTION

The past decade has seen an explosion in the amount of machine learning research employing deep learning techniques. Built on the foundation of learning task-specific, nonlinear, and increasingly abstract feature representations directly from raw data[1], these modern techniques have proven highly effective in a variety of domains, particularly those involving data that exhibits inherent local structure, such as spatial pixel relationships in computer vision[2,3] and sequential character-level or word-level associations in natural language processing[4,5]. Deep learning techniques have also been successfully applied in domains such as healthcare[6,7], finance[8], genomics[9], drug discovery[10], and speech recognition[11].

While the results from the deep learning revolution speak for themselves, a common limitation underpinning all deep learning methods in their purest form is the inherent difficulty or inability to precisely determine *why* a model arrives at a particular output or prediction. Along with fundamental properties of deep learning algorithms, this essence of explainability can be viewed as part of a two-sided coin: the hierarchical nonlinearities present in deep learning methods that hinder natural explanation of its processes are precisely why these models are so effective at developing high-dimensional and latent representations of raw data that yield superior results over more human-understandable methods. Often referred to broadly as *interpretability*, transparent explanatory processes for a model's inner workings are missing from fundamental deep learning algorithms and have led to their reputation as "black box" methodologies.

Although interpretability is widely viewed as a global limitation and a necessary element of future deep learning research, at current time there is still little consensus on a precise, formal definition of interpretability[12–14]. In a unique effort to ground the discussion, Lipton[12] outlined both objectives and properties of interpretable models, in which associated methods were categorized as either improving transparency, i.e. shedding light on a model's inner workings, or providing post-hoc interpretations, where explainable information is gleaned from already-trained models. In this paper, we build on Lipton's groundwork and utilize this taxonomy for discussing modern approaches for deep learning interpretability.

Given the widely publicized and overwhelmingly successful results of deep learning approaches for computer vision tasks, it comes as no surprise that the majority of interpretability research has also focused on the prevalent domain of image processing. Several works have reviewed interpretable methods for computer vision that capitalize on the spatial aspect of pixel-based image data[15–17]. Contrarily, in this paper we focus on methods for explaining and understanding *sequential* deep learning models that use sequential data inputs such as text, time series, and other data types that imply a notion of *order* or *time*. While some technical methodologies for processing sequential data overlap with those used for images, there is a distinct difference in application and interpretation, and in this paper, we highlight unique aspects of sequential understanding as it relates to deep learning frameworks.

Sequential data can be viewed as an ordered collection of samples $S = \{S_1, S_2, S_3, …, S_N\}$, where $N$ refers to the total number of instances, and each item $S_i = \{x_i^1, x_i^2, …, x_i^D\}$ refers to a single sequential instance comprised of $D$ features, where $D$ can be one or many values per time step. For real-valued time series such as stock prices or clinical vital signs, the local structure imposed within each sample is based on time. Alternatively, for textual data types the structure is based on the ordering of distinct characters or words that naturally carry meaning based on our understanding of language.

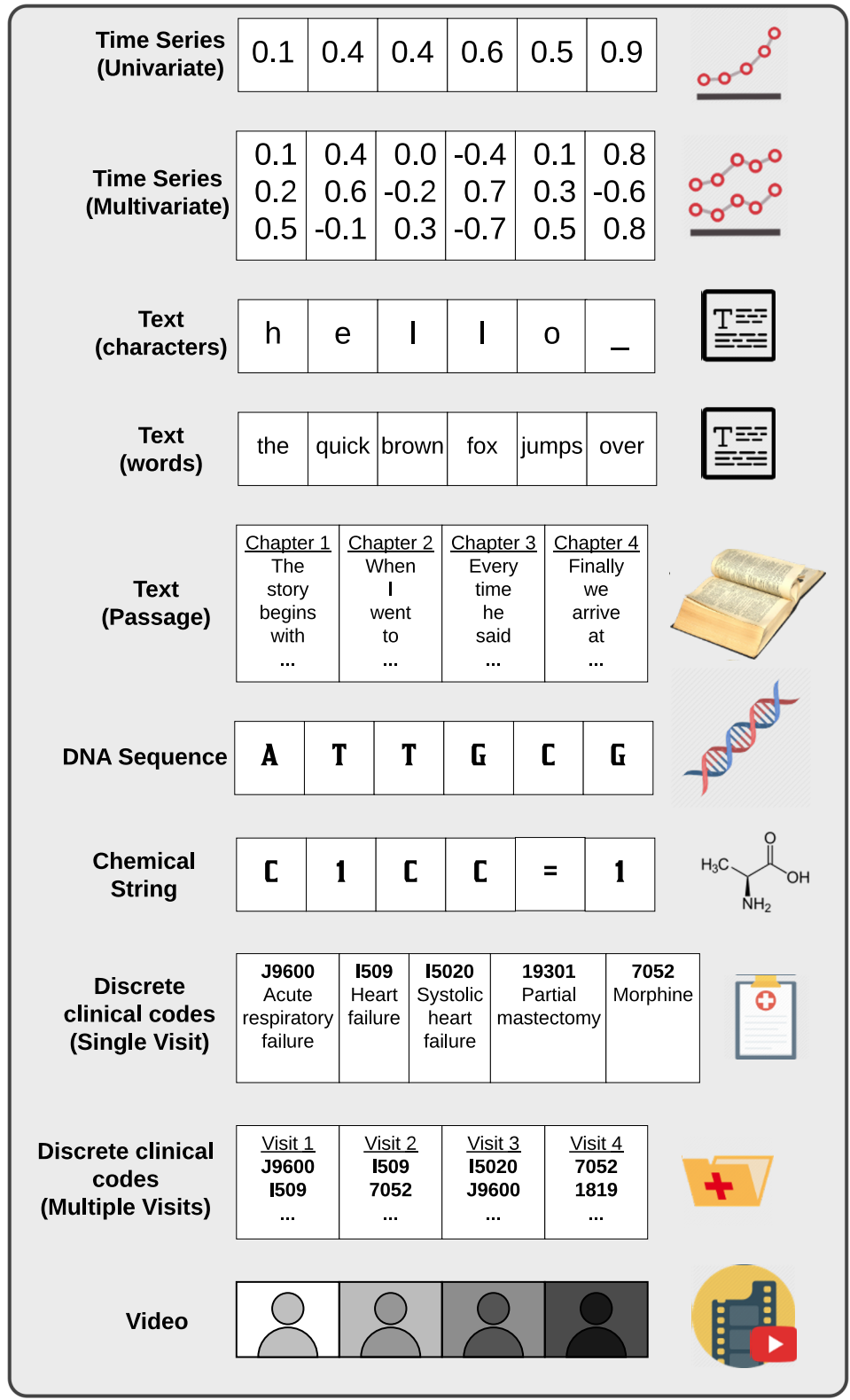

Figure 1. Brief overview of sequential data types used by deep learning studies included in this review.

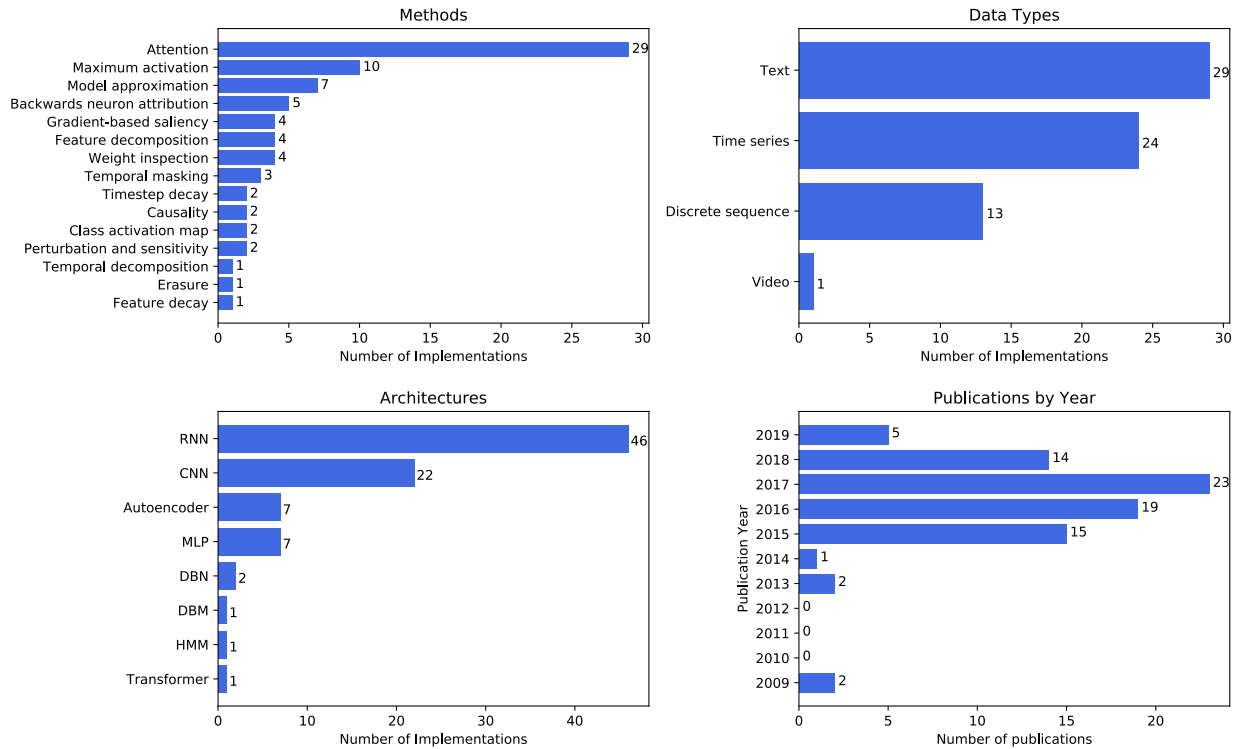

*Figure 2. Histograms of techniques, data types, architectures, and publication count by year for deep sequential interpretability studies included in this review.*

In the simplest case, sequential data is comprised of a list of single elements with corresponding time stamps. However, sequential data can also be multivariate; that is, it can be comprised of multiple values at each time step.

In this review, we provide an overview of published research involving the interpretability of deep learning methods in the context of processing sequential data types. Along with describing the methodology and contribution of each technique, we place all studies in context by deriving a goal-oriented taxonomy and categorization of techniques based on the nature of the interpretable information provided to practitioners. We conclude with a discussion of current trends and similarities, limitations, and future direction of deep sequential interpretability research.

## II. METHODS FOR DEEP SEQUENTIAL INTERPRETABILITY

There is a lack of consensus on a clear definition of interpretability, not solely for sequential data and applications, but across the entire field of deep learning. In fact, the notions of transparency, explainability, and the broad concept of *interpretability* can refer to several different aspects of algorithmic modeling. The desired outcome is often a human-understandable insight into the input-output data relationship, obtained from an opaque but effective sequential deep learning model. The specific types of interpretability techniques and information gleaned from them are widely varied and task, model, and data dependent. For example, a number of techniques identify important input features (Figure 2).

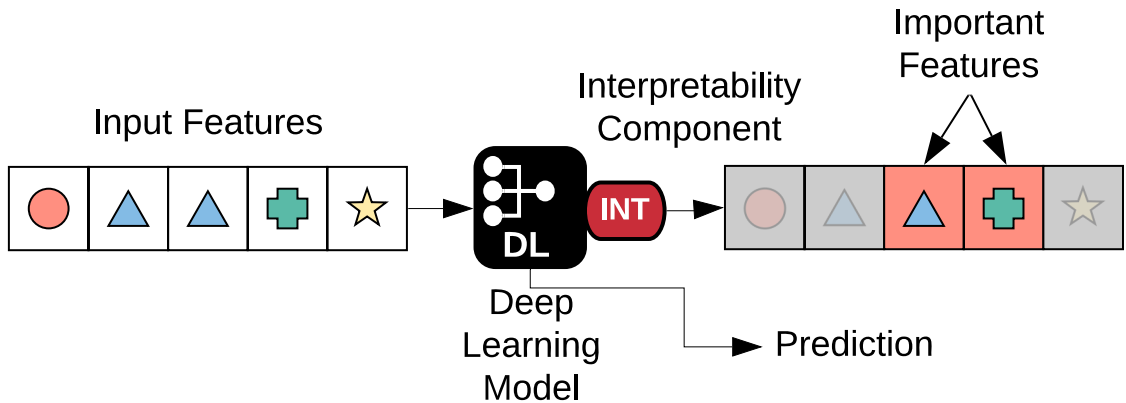

*Figure 3. Interpretability models can potentially identify important input features.*

We guide our overview of deep learning interpretability techniques for sequential data based on a categorization of their human-understandable interpretations. Based on a review of deep learning interpretability research over the past decade, we have identified three primary trends in the application of interpretable techniques to sequential modeling: (1) network analysis information, (2) sequential saliency, and (3) multivariate attribution information (Figure 2). For techniques that fall under multiple categories, we include them separately and discuss technical aspects relevant to the category of interest.

## A. Network Analysis

Network analysis techniques shed light on the precise mechanisms and data transformations that occur within a deep sequential neural network. These techniques examine individual neurons, layers, or regions of a trained deep learning model, and characterize the types of inputs that each have learned to identify. We also include class prototypes in this category, where synthetic inputs are generated that are most characteristic of a given class in a supervised learning setting – which can be viewed as the analysis of the final layer of a trained deep sequential model. All methods in this section operate on a trained deep model, and do not require modification of the underlying architecture.

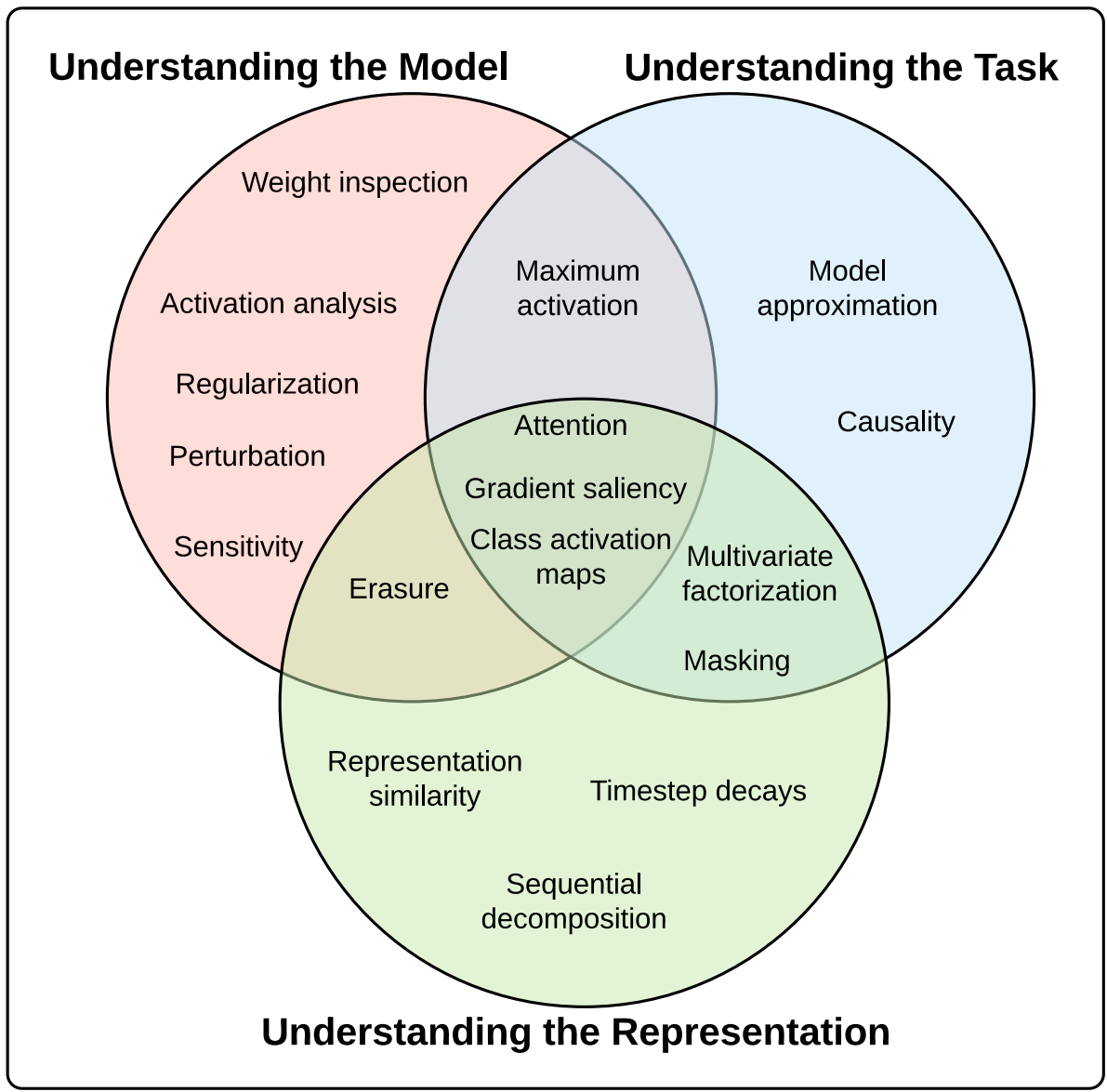

*Figure 4. Relationships between specific sequential interpretability techniques and target of increased understanding.*

***i. Weight inspection***
Perhaps the simplest technique to gain insight into a fully trained deep learning model is by directly examining its learned weights. These approaches typically constrain analysis to the weights of the first layer of a deep network where there is a direct interaction on the raw data inputs. We can inspect the weights in a regular fully connected network (Fig.1.a), in filters of a convolutional neural network (CNN, Fig.1.b), or in weights of a word embedding network (Fig.1.c).

This approach is common as a baseline step in image processing applications where weights of a trained CNN can be visualized and interpreted as edge, line, or shape detectors[21]. Karpathy et al.[19] extended this technique to analyzing sequences of image frames in their multiresolution CNNs for video classification, where they found qualitative differences between the types of colors and frequencies included in filter responses.

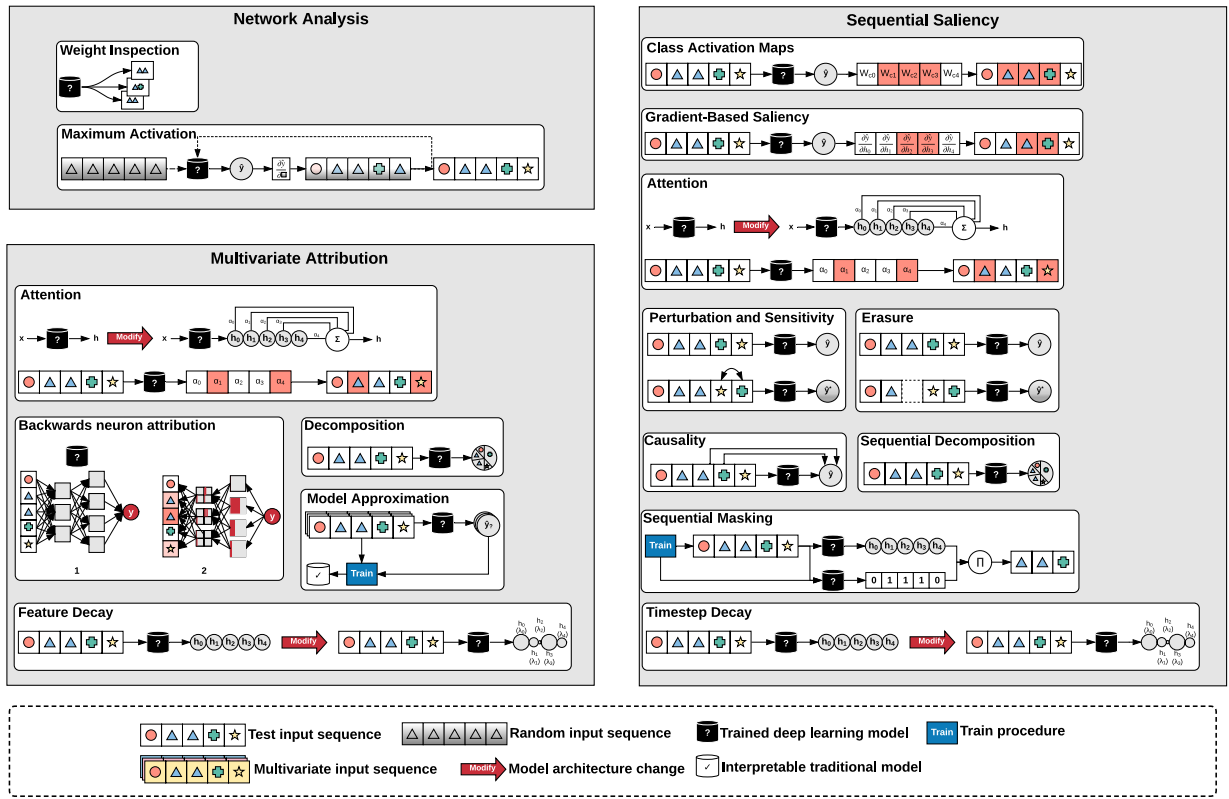

*Figure 5. High-level operational overview of sequential interpretability methods in a supervised learning setting.*

In the context of univariate time series, Lasko et al.[22] visualized the weights of a trained autoencoder for classifying uric acid sequences as exhibiting either gout or leukemia, where they found the first layer weights detected functional elements such as uphill and downhill ramps and other repeated motifs.

Mehrabi et al.[18] implemented a Deep Boltzmann Machine for analyzing sequences of two sets of medical codes, International Classification of Diseases (ICD-9) and Healthcare Cost and Utilization Project (HCUP), to identify common sequential patterns among patients with similar diagnoses. In a visualization heatmap of the first layer's hidden weights, they found commonalities between codes and identified trends relating to chronic disease.

Li et al.[20] visualized word and phrase embedding representations as both heatmaps and via t-SNE, finding particular dimensions of the learned representations corresponded to various syntactic and semantic aspects, and found qualitative clustering of locally compositional phrase types such as negation.

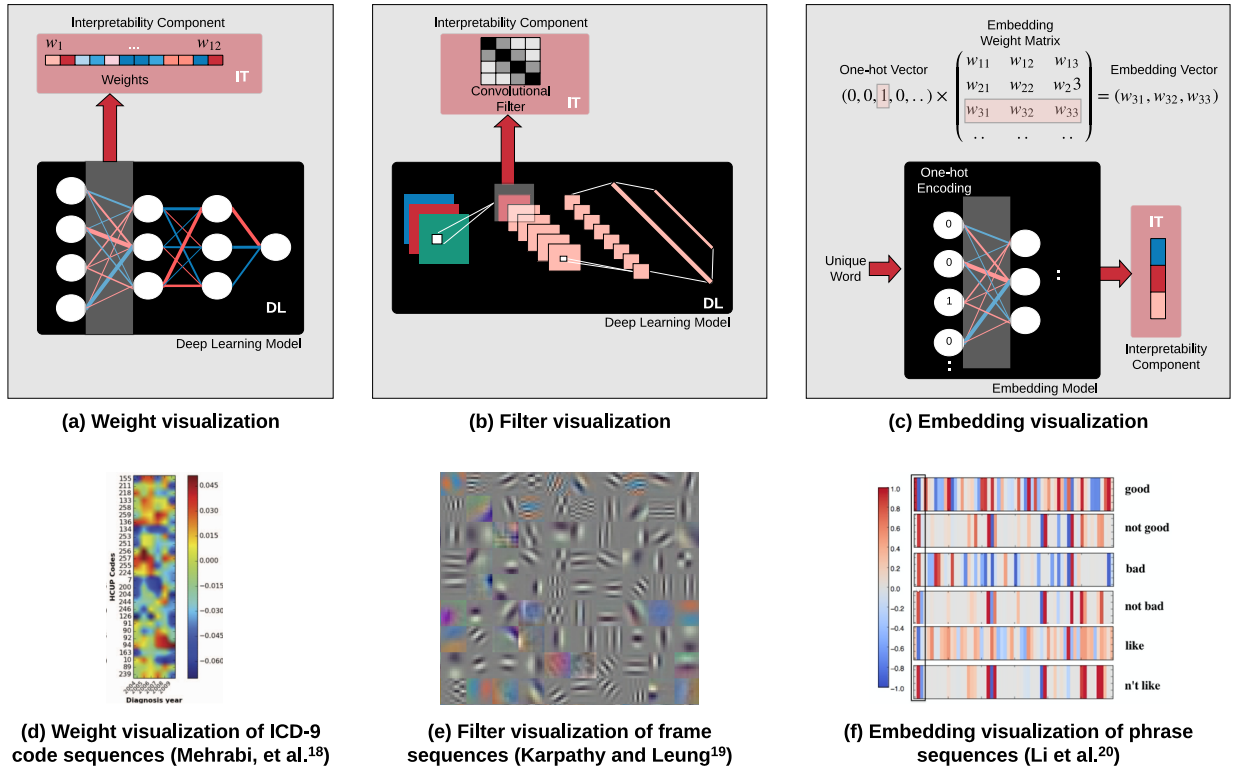

(a) Weight visualization  (b) Filter visualization  (c) Embedding visualization

(d) Weight visualization of ICD-9 code sequences (Mehrabi, et al.[18])  (e) Filter visualization of frame sequences (Karpathy and Leung[19])  (f) Embedding visualization of phrase sequences (Li et al.[20])

*Figure 6. As a baseline Interpretability Technique (IT), weight inspection methods inspect the weights, e.g. this can be done by visualizing (a) the weights in a fully connected network, (b) the filters in a CNN, or as (c) the embedding vector of a word embedding network  Example Applications of each : (d) Medical code sequence: weight visualization of the first hidden layer with [ICD9 x Diagnosis year] matrices as input[18], (e) sequence of video frames: filter visualization of video streams [19], (f) sequence of words: embedding visualization of sentiment in natural language[20].*

## *ii. Maximum activation*

Similar to visualizing fixed network weights of a trained model, the network activations within a deep model can also be examined. Unlike weight analysis, which is usually restricted to the first layer, activations at any stage of a deep network can be visualized along with their driving inputs. For each unit of a network, particular inputs can be determined – whether present in real data or generated – that maximally activate that unit[23], providing qualitative understanding of the types of inputs each portion of a network is looking for. It should be noted that individual neuron activation might not be very meaningful. As previously shown, it is a subset of neurons, rather than the individual units, that contain the semantic information in the high layers of neural networks.[24]

Hermans and Schrauwen[25] and Karpathy et al.[26] both take this approach for character-level recurrent neural networks, where cell activations were aligned with the inputs with highest activations to discover interpretable textual patterns such as quoted text and longer-range character interactions.

Dong et al.[27] analyzed driving styles using spatiosequential GPS time series using recurrent neural networks and visualized the activations of select neurons with maximum

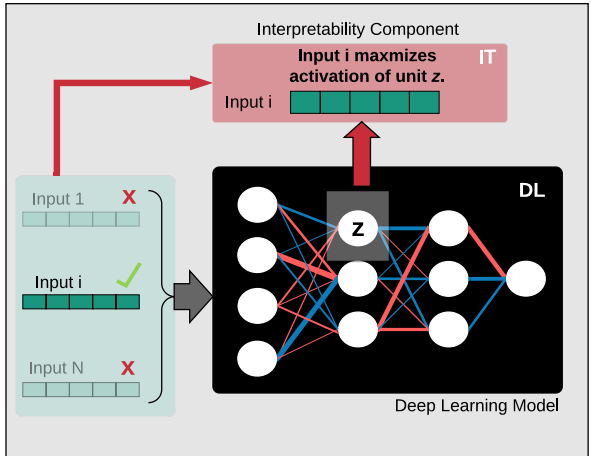 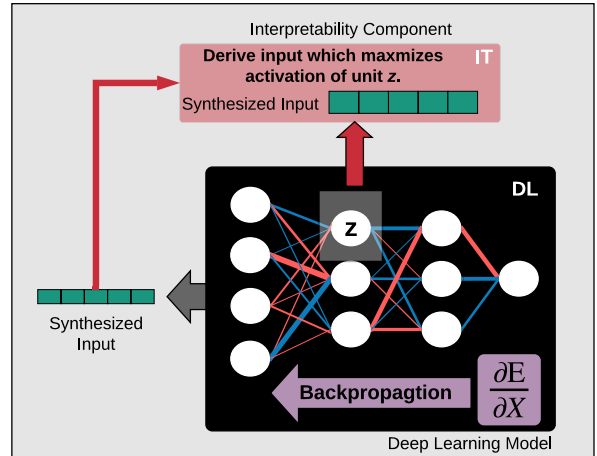

(a) Maximum activation using provided inputs.

(b) Maximum activation using generated input through backpropagation.

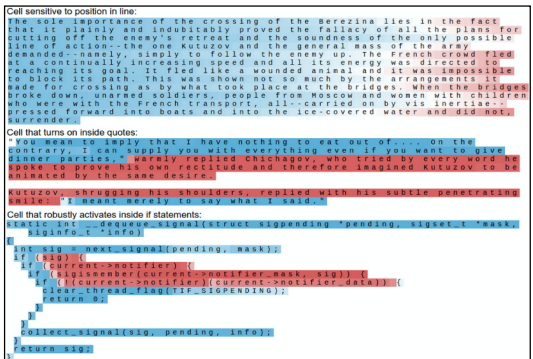 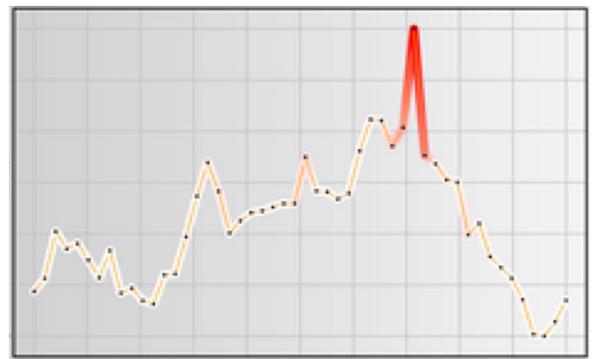

(c) Activation of an LSTM cell as processing the sequence of characters in a text passage. Text color corresponds to tanh(c), where -1 is red and +1 is blue.[24]

(d) The influence values of a time series are obtained by computing the gradient w.r.t. the input.[25]

Figure 7. For each unit of a network, particular inputs can be found, whether (a) present in real data or (b) generated. Examples: (c) Activation of an LSTM cell as processing the sequence of characters in a text passage, (d) The influence values of a time series.

activations, showing that particular neurons react to patterns such as angular trends, speed changes, and potential GPS failures.

Che et al.[28] and Kale et al.[29] utilized autoencoders and fully-connected networks along with prior-based regularization and causal associations for predicting disease from multivariate time series, in which physiological patterns from inputs that maximally activated hidden units were visualized to discover sequential phenotypes for individual diseases.

In the aforementioned studies, interpretations were based on the real dataset inputs that resulted in a unit's largest activation. However, gradient-based techniques can also be used to generate a synthetic input that maximally activates a given neuron[23]. This approach was taken by Lanchantin et al.[30,31], who applied convolutional neural networks for extracting motifs from genomic sequences. Once their model was trained, they optimized each class output with respect to an input sequence, and through backpropagation were able to generate class-specific motifs for understanding how their models were predicting transcription factor binding sites. Siddiqui et al.[32] take a similar approach in the context of univariate and multivariate time series.

## B. Sequential Saliency

The following techniques are designed to visualize the timesteps or sub-patterns of an input sequence that most contributed to the overall sequential representation used for final model prediction in a supervised learning environment. Application of these methods can be used to understand why a model has made a particular prediction by pointing back to specific elements of the input sequence. In contrast to the previous section, the following methods are focused more on characterizing the relationship between input time steps and task labels ("*why* was the prediction made?"), and are less concerned with determining the internal mechanisms by which this happens ("*how* was the prediction made?").

### *i. Class activation maps*

Wang et al.[33] utilized fully convolutional networks for a variety of univariate and multivariate time series classification tasks. They used a class activation map to assign a saliency score to each element of the input time series, where the activations of a single time step are summed over all filters in the penultimate convolutional layer. A class activation map can be computed for each available class as shown below:

$$Class\ Activation\ Map = \sum_{k} w_k^C A_k$$

By superimposing saliency scores along with the original sequence, they were able to highlight time series segments that were most aligned with each available classification target.

### *ii. Gradient-based saliency*

In addition to previously mentioned maximum activation interpretations based on optimizing an input sequence with respect to a given class prediction, Siddiqui et al.[32] also visualized the importance of each of a series of convolutional filters by computing the gradient of each filter's output with respect to the input sequence. Similarly, Lanchantin et al.[30,31] compute the gradient of the output layer with respect to the input layer to determine the relative importance of each element of genomic sequences.

In general, we can compute a score function by computing the first order Taylor expansion.[30] We can then compute the derivative using backpropagation in the network. A point-wise multiplication of the saliency map with the input features will provide the influence of each individual feature in the sequence on the output score.

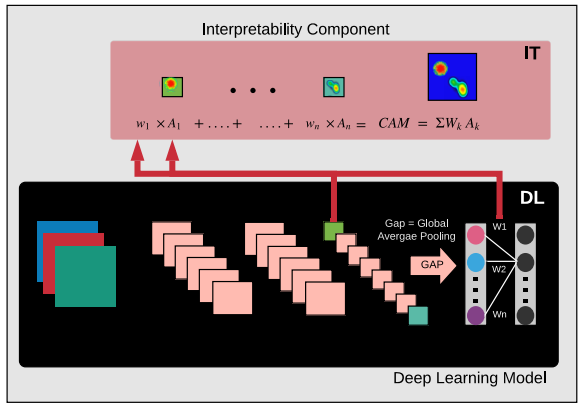
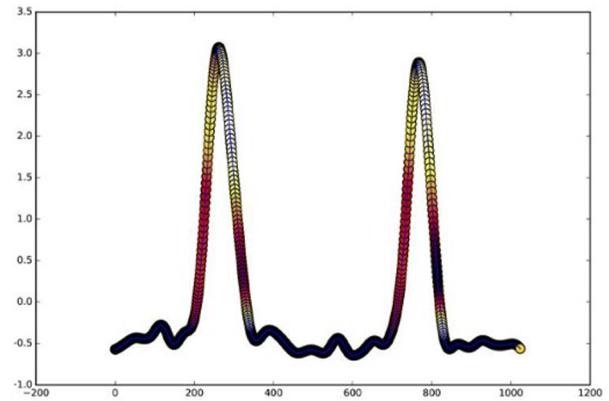

(a) Class activation map

(b) Class activation map on 'StarLightCurve' dataset. Discriminative regions of the time series for the right classes are highlighted.[32]

*Figure 8. (a) The class activation map technique multiplies weights $w_i$ by activation maps $A_k$ to generate the activation maps. Global average pooling is used to summarize each activation map of the final convolutional layer. Class activation map is applied on a time series dataset to highlight discriminative segments. (b) Example of applying class activation maps to highlight salient timesteps of a univariate time series.*

Unlike the maximum activation approach, this approach does not examine activation of individual hidden units, rather it examines saliency of the input features.

$$S(X) \approx w^T X + b$$
$$Saliency\ Map = \frac{\partial S}{\partial X}$$

Wang et al.[34] also took this approach, where their multilevel Wavelet Decomposition Network, which hierarchically decomposed a time series into low and high-frequency signals in the time-frequency domain across several stages, utilized a similar gradient-based derivation of important elements of both the input and intermediate layers for a variety of time series classification and forecasting tasks.

### *iii. Attention*
Perhaps the most widespread form of sequential interpretability in recent years, attention mechanisms have found their way into a variety of models and applications.

Conceptually, attention mechanisms are designed to allow a model to focus on smaller subsets of an input that are most influential with respect to an output task. Examining how a model has decided to focus its attention weights for a given input allows improved interpretability and explanation towards a model's overall prediction.

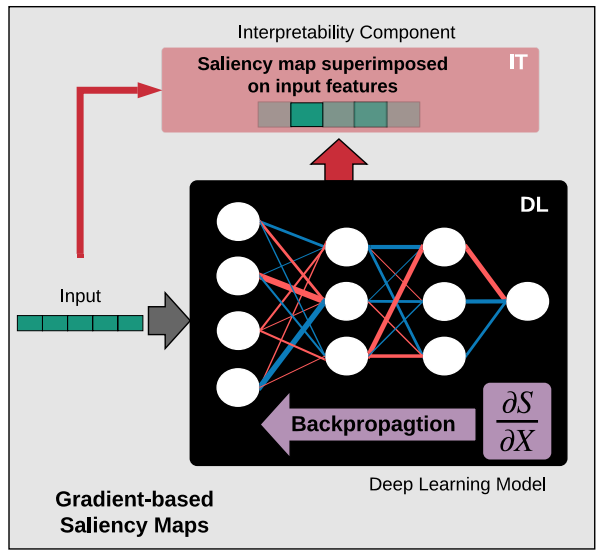
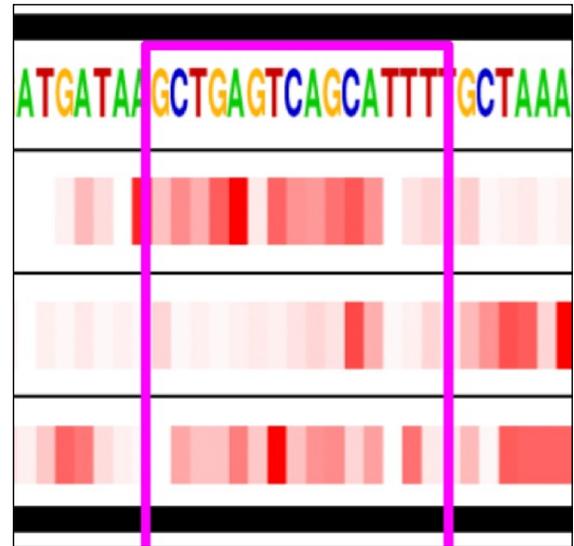

(a) Gradient-based saliency

(b) Saliency maps of a DNA-base sequence. Each sequence consists of 101 DNA-base characters (A,C,G,T).[30]

*Figure 9. (a) Gradient-based saliency map examine saliency of individual features by superimposing the saliency map on the features. (b) Example using saliency maps to identify important DNA-base characters of genetic sequences.*

Attention mechanisms can take several forms, but all typically operate on hidden state vectors at an intermediate or penultimate layer in a deep neural network. For facilitating discussion, suppose we have an input sequence $x$ of length $T$, with corresponding hidden state vector $h$ also of length $T$.

$$\alpha_t = \frac{e^{(W \cdot h_t)}}{\sum_{i=0}^{T} e^{(W \cdot h_i)}}$$

In the above example, attention – which is normalized to sum to 1 – is distributed to every element in a sequence based on each time step's compatibility with a global learned attention matrix $W$, evaluated here via dot product. The sequence can then be weighted by each time step's attention weight, and in some sequential tasks, an overall representation of the sequence (often referred to as a context vector) is computed by summing over all time steps as opposed to taking the final time step representation as would be done without attention:

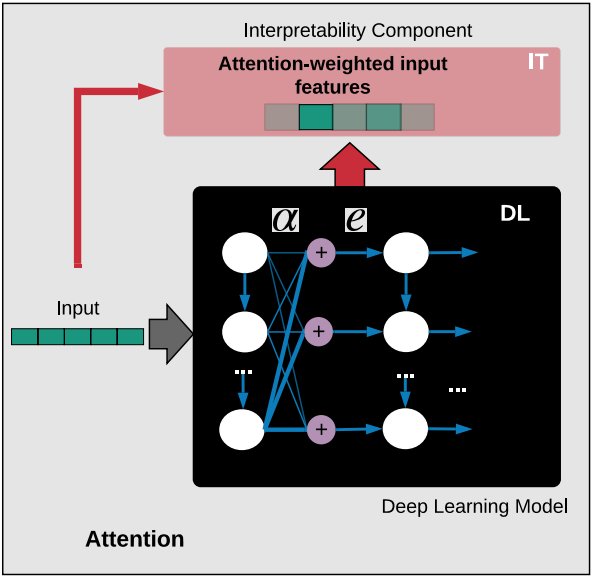 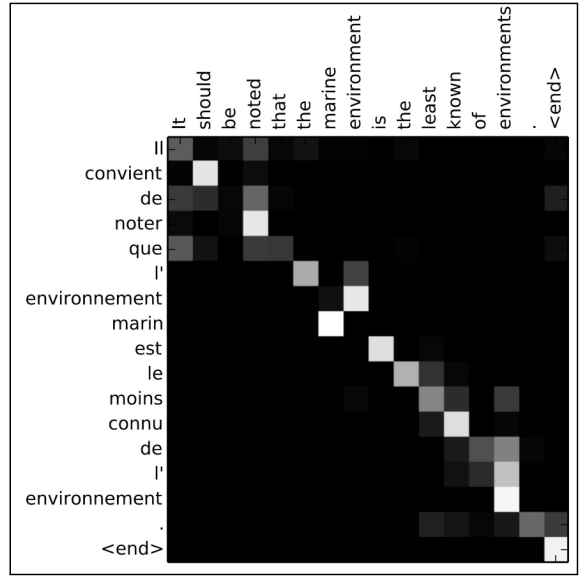

(a) The attention mechanism computes a context vector by summing over all time steps as opposed to taking the final time step representation.

(b) The x-axis and y-axis correspond to the words in the source sentence (English) and the generated translation (French), respectively. Each pixel shows the weight of the annotation of the j-th source word for the i-th targetword.[37]

*Figure 10. (a) Attention-based mechanism (b) Example using attention to discover relative word weights between sentence pairs in two languages for a machine translation task.*

$$h = \sum_{i=0}^{T} \alpha_i h_i$$

Earliest use of attention mechanisms can be traced to the machine translation work by Bahdanau, et al.[35], whose attention-based method closely resembles the above equations in the context of their RNN-based encoder-decoder architecture for sequence-to-sequence translation. Attention scores over word representations in the source language were computed when generating each successive word in the target language, and were based on an alignment score from a feedforward neural network. Prior to the use of attention mechanisms for machine translation, word decoding typically used the final recurrent token representation as the source context vector. This had the potential to "forget" very early words. When using attention, a context vector could instead be derived using a weighted sum of every input token.

Extensions of this early work for machine translation include Luong et al.'s notion of *local* vs. *global* attention[36]. For each decoding step in their model, global attention attended to every word in the source sentence in a similar spirit to Bahdanau et al[35]; local attention instead assigned word-level alignment scores within a fixed-size window surrounding a predicted position. The concepts of local and global attention exhibit parallels to the notions of *soft* vs. *hard* attention from the image captioning work of Xu et al.[37], in which soft attention generated differentiable alignments over all input elements, whereas hard attention selected specific patches of the input before attending in a nondifferentiable manner. The work of Luong et al. additionally proposed alternative alignment functions for assessing representation compatibility, including dot product and forms of general attention based on learned weight matrices.[36]

In the context of sequential deep learning, these early attention mechanisms allowed models to piecewise utilize the hidden representation of every individual element from an input sequence, rather than operating solely on the aggregated context vector from the final hidden state representation of a recurrent neural network. One perspective of sequential attention can be seen as exposing the entire history of the input representations to the model, in effect yielding a notion of internal memory[38]. In the implementation of memory networks for the task of question-answering by Sukhbaatar et al.[39], an input sequence's tokens were individually and simultaneously embedded into memory vectors. Compatibility between each input token's memory vector and a given query vector was computed using an inner product and was used to weight the input sequence into a single context vector as described in previous attention works. As sequence elements were individually embedded, relative order was preserved by adding a positional encoding to each element based on a learned embedding of its relative index in the sequence. For the task of question answering, their model involved recurrent hops across an external memory. Given the state of modern attention mechanisms, the design of this memory network was notable for its elimination of recurrent processing of sequential inputs, and for the use of a learned position embedding to account for relative sequential ordering.

Utilizing a similar style of non-recurrent processing of sequential inputs, Vaswani et al.[40] introduced the attention-based Transformer model for the task of machine translation, which outperformed traditional recurrent encoder-decoder paradigms. The Transformer viewed attention as a function of keys *K*, queries *Q*, and values *V*. In their work, all three elements came from the same input sequence, and is why their style of attention is referred to as *self-attention*. In a similar manner to previously described works, compatibility between a key and query is used to weight the value, and in the case of self-attention, each element of an input sequence is represented as a contextual sum of the alignment between itself and every other element. Similar to the memory networks of Sukhbaatar et al.[39], the Transformer also involves the addition of a positional encoding vector to preserve relative order information between input tokens. The recent NLP method BERT[41] is based on Transformers and at present time whose variants represent state of the art in a variety of natural language processing tasks.

Given the brief evolution of attention mechanisms above, it is no surprise that various forms of attention have seen most widespread use in a variety of NLP-adjacent applications, such as machine translation[35,36,40,42–44], sentiment analysis[45–47], text entailment[45,47,48], question answering[49], text summarization[50], recommender systems[51], image captioning[37], and visual question answering[52–54].

As attention mechanisms evolved primarily for improved NLP task performance, other studies noted that such methods could be used as a window into the most impactful input sequence elements contributing to a contextual representation for classification. This perspective on attention is a form of sequential saliency, and can be applied to many types of sequential deep learning aside from natural language processing applications.

Chorowski et al.[55] implemented attention in a sequence-to-sequence framework for speech recognition, where a context of attended time steps improved the ability to generate precise phoneme locations from audio recordings.

Several works have implemented attention mechanisms for the prediction of clinical outcomes from sequences of discrete medical codes. In their GRAM model for electronic health record prediction tasks, Choi et al.[56] constructed a directed acyclic graph from medical ontologies and used attention to select relevant graph segments for predicting diagnoses with a recurrent neural network from sequences of discrete clinical codes.

Ma et al.[57]'s Dipole framework utilized attention with bidirectional RNNs for modeling longitudinal hospital visits.

Sha and Wang[58] implemented a hierarchical attention mechanism that focused on relevant visit-level and code-level representations in a GRU for predicting hospital mortality. A similar approach was taken by Zhang et al.[59] in their Patient2Vec system for predicting hospital readmission risk.

Lin et al.[60]'s HA-TCN model stacked several sequential convolutional layers with increasing dilation, and utilized both within-layer and across-layer attention mechanisms for predicting myotonic dystrophy diagnosis to capture patterns of variable scale.

In a general warning, Pruthi et al.[61] outlined two primary drawbacks of relying on token-level attention scores as a means of model interpretability or quantifying the most predictive input features. In their work, they illustrate how attention scores can be easily manipulated by small modifications to the objective function. They additionally highlight the issues of attending to positional hidden representations, which contain information from neighboring words in the input sequence. Their experiments showed that even when preventing the assignment of attention to crucial words in a sentence, original performance does not suffer.

### *iv. Perturbation and sensitivity*

While many aforementioned methods have utilized weights and activations of a trained network to derive sequential explanations, Alvarez-Melis and Jaakkola[62] generate input-output token dependencies for a sequence-to-sequence framework without using a trained model's parameters. They describe a multi-stage framework in which input sequences are perturbed using a pretrained variational autoencoder to generate semantically similar sentences, and for each token in the target sequence, a probabilistic model is estimated to learn dependency coefficients between the original input and target output, taking the form of a dense bipartite dependency graph. In their final explanation

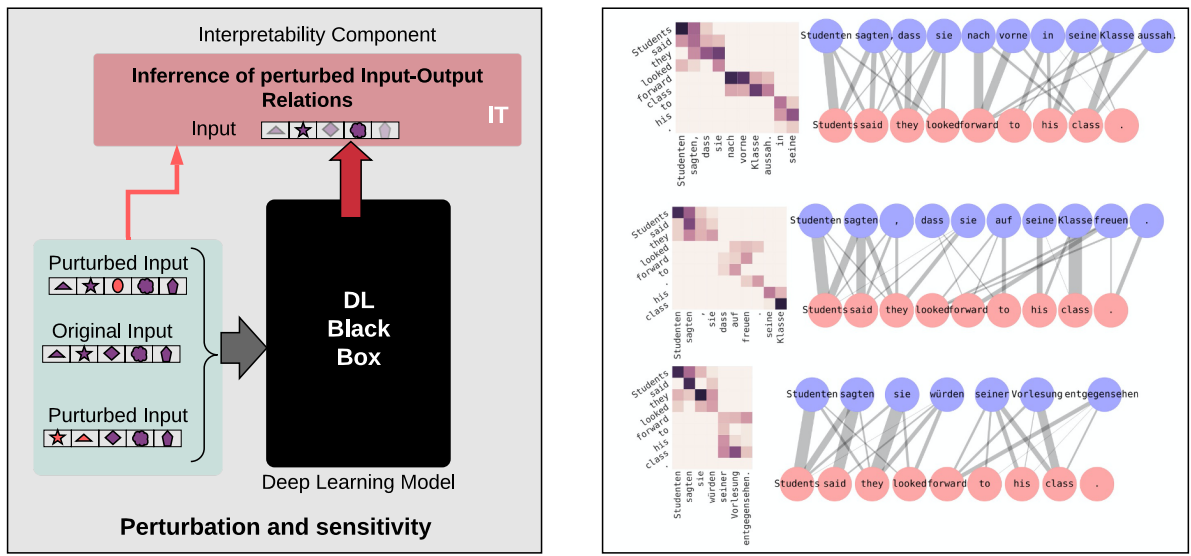

(a) Input sequences are perturbed to learn dependency between the original input and target output.

(b) Explanations for the predictions of three Black-Box translators: Azure (top), NMT (middle) and human (bottom)[57]

*Figure 11.* (a) Intermediate or final model outputs of perturbed sequences can be examined to discover influential elements in a prediction task. (b) Example on a machine translation task.

step, they apply graph partitioning algorithms that incorporate dependency uncertainties to discover strong word dependency explanations.

Alipanahi et al.[9]'s DeepBind framework employed a type of sensitivity analysis in their mutation maps for predicting DNA and RNA binding protein specificities, where sequence elements were replaced by molecular mutations and overall predictions compared with that of the original input sequence.

### *v. Sequential masking*
In an effort to better explain the prediction of multi-aspect sentiment analysis and retrieval of semantically similar questions, Lei et al.[63] augmented their RCNN model[64] with a jointly trained probabilistic generator network that learned a binary vector over input words most influential for a model's prediction. Along with regularization terms to enforce short, contiguous phrases, they interpreted selected subsequences as an interpretable prediction rationale.

Choi et al.[65] implemented a form of masking for their context-dependent word embeddings, where a sigmoid-activated mask over sequence words was learned jointly before word embeddings were fed into an LSTM network for machine translation. They found their masking procedure allowed the translation model to focus on particular embedding dimensions of similar words whose meaning depended on surrounding context.

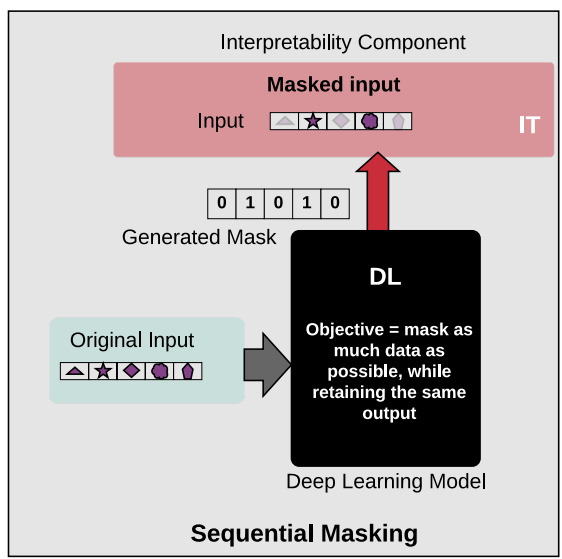
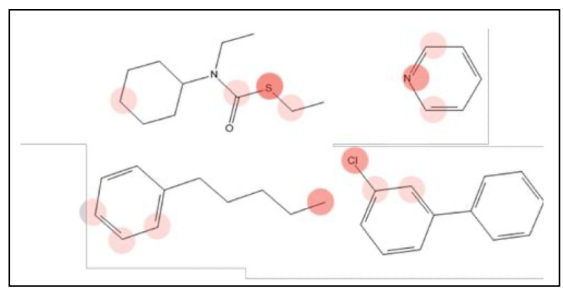

(b) Colored circles of increasing darkness indicate the locations of increasing attention on the molecule, obtained through masking input SMILES string by a network.[61]

(a) The objective is to mask as much data as possible, while retaining the same output.

Figure 12. (a) By learning to mask or ignore elements of input sequences and comparing outputs with unmasked models, one can reveal elements crucial to a given prediction. (b) Example using masking for chemical molecules represented as SMILES sequences.

Similarly, using an already-trained recurrent and convolutional model for predicting chemical properties from SMILES string sequences, Goh et al.[66] trained a separate residual network to learn a continuous sequential mask over input elements in an effort to mask as much input data as possible without changing the output of the trained model. When examining the learned masks, they found localized patterns that conformed to known chemistry concepts.

### *vi. Erasure*

Using the technique of representation erasure, Li et al.[67] introduced a general method for measuring the influence of particular words or phrases, word representation dimensions, and intermediate hidden units and layers in their framework applicable to a variety of NLP tasks. In their derivation, the log-likelihood of a trained model to predict the correct label $c$ for an input sequence $e \in E$ can be represented as $S(e, c) = -\log P(L_e = c)$, where $L_e$ is the model's predicted label. When analyzing a particular dimension of interest $d$ within a trained model, representations involving that dimension are set to zero, with the resulting prediction log-likelihood represented as $S(e, c, \neg d)$. The importance of dimension $d$ is defined by the relative change in log-likelihood over the entire corpus $E$ as shown below:

$$I(d) = \frac{1}{|E|} \sum_{e \in E} \frac{S(e, c) - S(e, c, \neg d)}{S(e, c)}$$

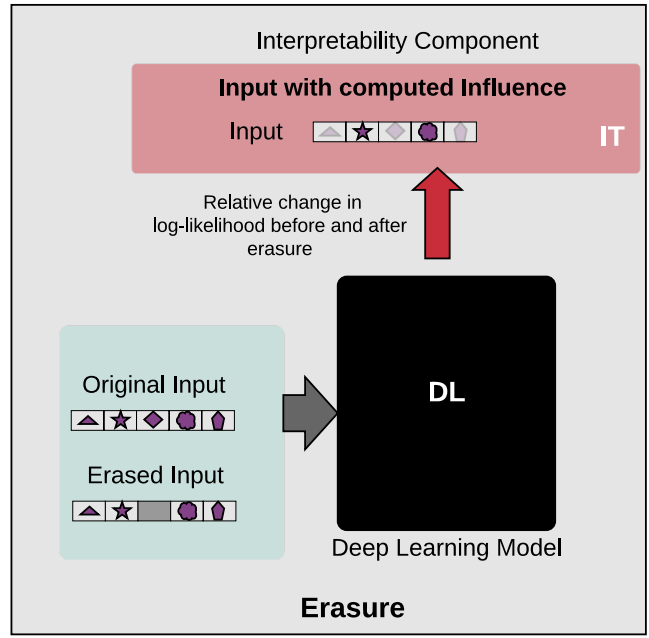

*Figure 13. Input elements or hidden layer representations can be set to zero to measure their relative impact. The importance of the erased dimension is defined by the relative change in log-likelihood over the entire corpus.*

The method of representation erasure yielded important words, embedding dimensions, and hidden units for a variety of NLP tasks including sequence tagging, ontological classification, and sentiment analysis. They also explored reinforcement learning for determining the minimum number of removed input words to change their model's prediction decision.

### *vii. Causality*
Kale et al.[29] employ the Pairwise LiNGAM method of causal inference to deep representations of physiological time series to discover the most causal latent dimensions of autoencoder-based representations. For the most causal hidden features, they visualized the mean of the select input time series that resulted in maximum activation of the most causal units and interpreted causal clinical phenotypes for two healthcare prediction tasks.

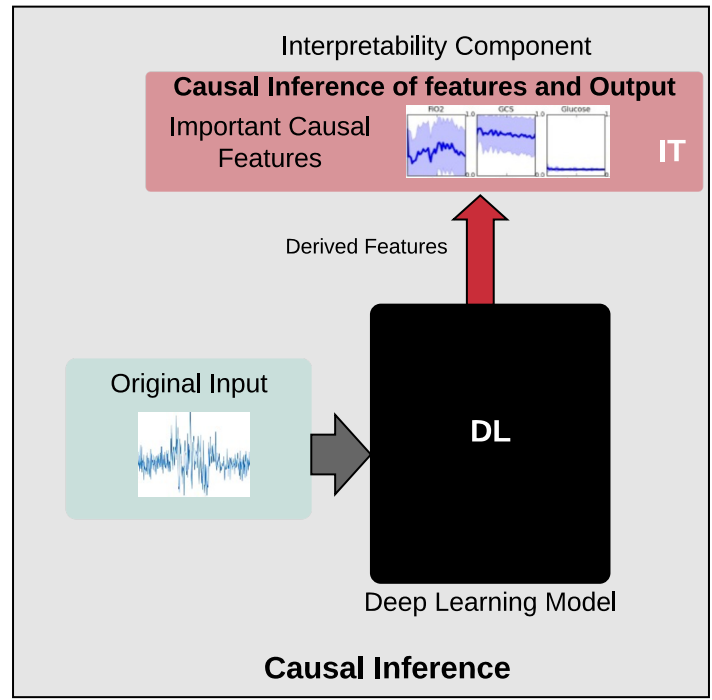

*Figure 14. Important features with respect to the output can be identified using causal inference.* [29]

### viii. Sequential decomposition
Murdoch et al.[68] derived a method to decompose the output and cell states of LSTMs to isolate the contribution of a current phrase from the remainder of a textual sequence. Qualitative comparison with several related studies[67,69,70] yielded more sensible prediction rationale for a few select examples in sentiment analysis tasks, and when comparing the distribution of positive and negative phrases, they found their method of contextual decomposition to produce more separated sentiment densities.

### ix. Timestep decay
For retrieving semantically related questions in a question-answering forum, Lei et al.[64] modified a convolutional neural network with a context-dependent gating scheme, similar in spirit to an LSTM, but involving a decay term $\lambda$ for adaptively ignoring irrelevant textual tokens. When visualizing a heatmap of word-level decay terms for several dataset examples, they found increased focus on domain-specific words important for their retrieval task.

## C. Multivariate Attribution
Many of the aforementioned studies have focused on inputs of a single dimension, such as univariate real-valued time series or natural language text. An additional form of sequential interpretability arises when considering multivariate inputs, where at each time step only a subset of variables are relevant. Similar to sequential saliency, methods for multivariate attribution often take the form of visualization and pointing to specific segments of input sequences.

### *i. Attention*

The RETAIN framework by Choi et al.[71,72] made interpretability a primary goal, decoupling the standard attention mechanism from time step representations to derive a precise formulation of the contribution of each time step and feature dimension towards predicting heart failure from sequences of discrete clinical codes. In their work, a separate attention mechanism was used for assessing the importance of each sequential timestep, as well as each variable at a given timestep.

Xu et al.[73]'s RAIM model took a similar approach to distinguish between time and feature-based attention mechanisms in their guided attention method. Their model was used for predicting decompensation and length of stay using both discrete clinical codes and physiological time series in the intensive care unit. They also used two distinct attention mechanisms: one each for timestep and variable-level importance.

Similar to RETAIN[71] and RAIM[73], Qin et al.[74] implemented dual attention mechanisms along with LSTMs for forecasting real-valued financial time series that focused on both the timestep and the feature dimension of input multivariate sequences. The two attention mechanisms were designed to highlight both timestep and variable-level importance. Similarly, the GeoMAN framework of Liang et al.[75] incorporated both spatial and sequential attention mechanisms for forecasting geo-sensory time series.

Shickel et al.[76] implemented a self-attention mechanism in conjunction with a GRU network for the real-time prediction of intensive care unit mortality based on a collection of physiological signals, and mapped attention scores back to original time series inputs to provide interpretable explanations of each input variable at each time step to clinical practitioners.

### *ii. Backwards neuron attribution*

Introduced by Bach et al.[77] in their image processing work with the goal of quantifying the contribution of each input pixel towards the final classification prediction, the technique of layer-wise relevance propagation (LRP) frames a model's output $f(x)$ as a sum of relevance scores $R$ over all dimensions of an input $x \in \mathbb{R}^D$. In the context of images, each dimension $d$ is a pixel:

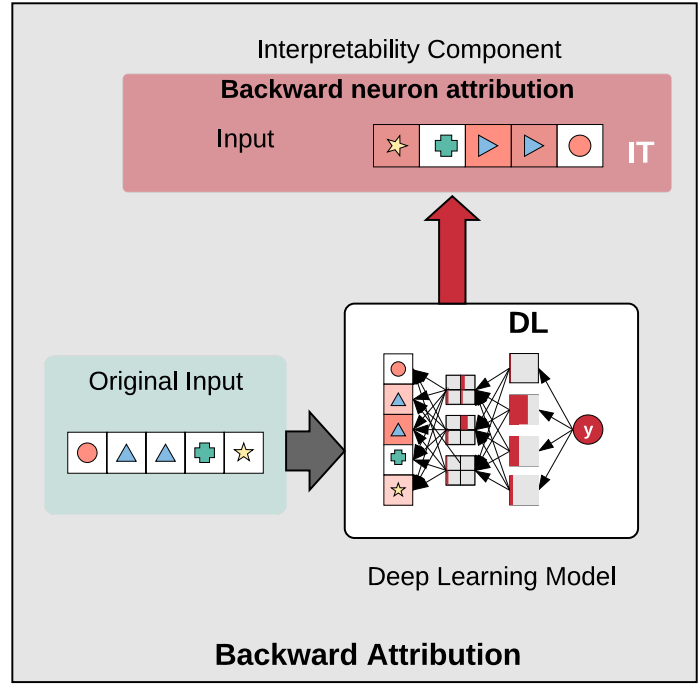

*Figure 15. The backward attribution methods distribute the relevance of an upper-level neuron to all lower connected neurons based on the relative neuron activations.*

$$f(x) \approx \sum_{d \in x} R_d$$

LRP operates by distributing neuron relevance at each layer $l$ of a network of $k$ layers such that the sum of a layer's neuron relevance scores $R_d$ is equivalent at each layer of a network, including at the input layer:

$$f(x) = \sum_{d \in l_k} R_d^{l_k} = \sum_{d \in l_{k-1}} R_d^{l_{k-1}} = \sum_{d \in l_{k-2}} R_d^{l_{k-2}} = \cdots = \sum_{d \in x} R_d$$

Layer relevance scores are computed in a backwards fashion, beginning with the quantity $f(x)$ of the output neuron in the case of binary classification. Overall, the relevance of a neuron in layer $l_k$ is determined by the relevance of each neuron in adjacent upper layer $l_{k+1}$ for which a connection exists between the two. An *a priori* function $R_{i \leftarrow j}^{l_k \leftarrow l_{k+1}}$ is used to quantify the relevance distributed to a neuron $i$ in lower layer $l_k$ from a neuron $j$ in upper layer $l_{k+1}$. The total relevance of a neuron $i$ in layer $l_k$ is computed by the sum of partial relevance scores passed from connected neurons in the following layer:

$$R_i^{l_k} = \sum_{j \in l_{k+1}} R_{i \leftarrow j}^{l_k \leftarrow l_{k+1}}$$

One potential form[77] of a relevance passing function is shown below, which distributes the relevance of an upper-level neuron to all lower connected neurons based on the relative neuron activations $a$:

$$R_{i \leftarrow j}^{l_k \leftarrow l_{k+1}} = R_j^{l_{k+1}} \cdot \frac{a_i w_{ik}}{\sum_h a_h w_{hk}}$$

Sturm et al.[78] applied LRP for understanding the decisions made by a fully-connected network for classifying EEG time series, and found LRP-derived relevant time series segments to align with neurophysiologically plausible patterns.

Along with gradient-based saliency techniques, Arras et al.[79] also used LRP to discover influential words for sentiment analysis tasks. Along with visualizing relevant words for single example documents and aggregating top words across an entire corpus, they also performed a quantitative experiment involving deletion of LRP-based influential words and found a significant performance decrease.

Similar in spirit to layer-wise relevance propagation, the DeepLIFT technique proposed by Shrikumar et al.[80] involves backwards neuron attribution assignments based on a unit's activation in relation to a reference value computed from a forward pass through the network. Caicedo-Torres and Gutierrez[81] implemented DeepLIFT in their multi-scale convolutional networks for predicting mortality in the intensive care unit from multivariate physiological time series, and found influential subsequences for single prediction as well as population-aggregated features corresponding to both positive and negative overall importance.

While not primarily used for multivariate attribution, we incude these techniques due to their inherent capability to do so. Backwards neuron attribution methods can also be used for highlighting sequential saliency.

### iii. Feature decomposition

With a distinct focus on sequential data, Hsu et al.[82] developed a factorized variational autoencoder model that isolates the global sequence-level and local segment-level attributes of an input sequence. They visualized qualitative impact of the disentangled, factorized attributes by varying the resulting latent dimensions for speech generation and evaluated quantitative results on speaker identification and phoneme recognition tasks.

Based on the work of He et al.[83], the MV-LSTM[84] and IMV-LSTM[85] architectures from Guo et al. refactor the traditional hidden state $h_t$ into a tensor of separate hidden states per input dimension $[h_t^1, ..., h_t^N]^T$. By deriving a set of complementary state update rules and aggregating context with a standard attention mechanism, the hidden states with respect to each input variable were isolated and visualized for several multivariate time series classification tasks.

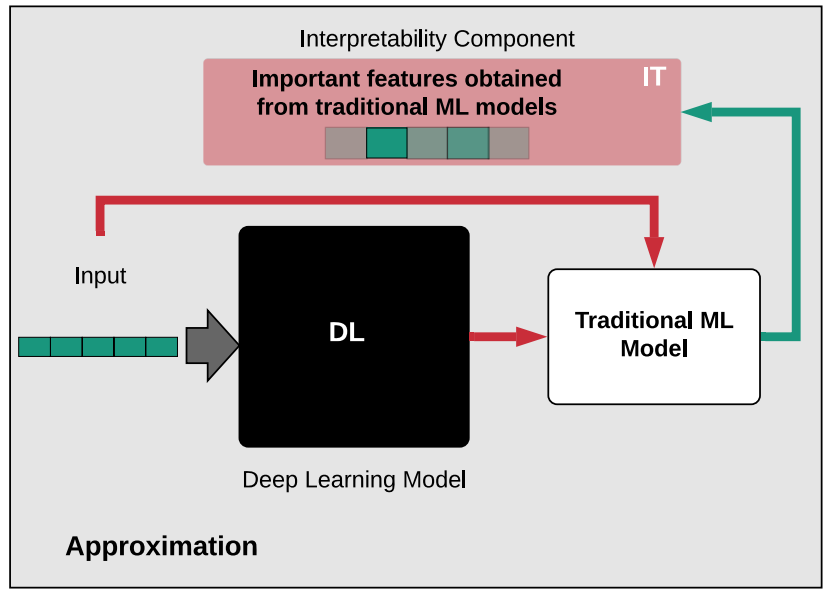

*Figure 16. A traditional machine learning model is trained on output from the deep learning model. The traditional machine learning model can highlight important features.*

### *iv. Feature decay*
Che et al.[86]'s GRU-D model included jointly trained, per-variable decay terms $\gamma$ for both input and hidden states for handling irregularly sampled and missing data in clinical time series. By examining the learned sequential decay vectors for each input variable, they were able to identify both the variables whose current value was important for predicting mortality, and the variables for which the model relied more on past observations.

### *v. Model approximation*
In the Interpretable Mimic Learning framework of Che et al.[87–89], a deep learning model such as DNN, SDA, or LSTM, was trained to predict health-related outcomes based on multivariate clinical time series. In one variant, a separate gradient boosted tree (GBT) model was trained on the original input time series to minimize the mean squared error between the true classification target and the raw output from the deep learning model. Another variant included an intermediate logistic regression classifier trained to predict the true target based on the raw deep learning outputs, in which the intermediate model's class predictions served as classification target for the GBT model based on the original input time series. In both cases, they were able to derive human-interpretable decision trees based on the learned knowledge from sequential deep learning.

Similarly, Wu et al.[90] developed a collection of regularization techniques involving the training of decision trees to predict the output of a GRU network given the original input time series. As with the work of Che et al.[87–89], explainable feature importance were available given the modeling of the deep network with an interpretable decision tree formulation.

In their LSTM framework for sentiment analysis and question answering, Murdoch and Szlam[69] derived a decomposition of an LSTM's output into a product of contribution factors for each word in an input sentence describing their influence towards the prediction of each class. They scored relevant phrases for each classification target across an entire corpus by measuring average contribution for each possible class in relation to all other classes. Following the training of the LSTM, they trained a simple pattern-matching model based on extracted relevant phrases for final prediction, and found competitive, although lower, results compared with the standard LSTM.

Krakovna and Doshi-Velez[91] trained hidden Markov models to approximate hidden states of an LSTM for character-level language modeling from text. They train decision trees to predict hidden states from their hybrid algorithm, allowing for insight on how the language model is constructed.

In their CNN framework for processing EEG signals, Schirrmeister et al.[92] correlate interpretable, known features such as mean frequency band envelopes with given units' activations within their receptive field. They analyzed the degree of correlation to interpret whether the CNN was approximating known predictive features or was learning representations that contained additional information. With their feature correlation maps, they discovered plausible localized patterns that supported existing research on the influence of EEG spectral power bands.

Similar to backwards neuron propagation, model approximation techniques are more generalized and can be used for both sequential saliency as well as multivariate attribution.

## D. Tools

Although this review has focused exclusively on interpreting deep learning methods in the context of sequential input data, we briefly note the presence of several open source tools and packages for improving general explainability for a variety of input data and model types. For Pytorch and Tensorflow, framework-specific tools with a suite of interpretability techniques include Facebook Captum[1] and tf-explain[2], respectively. Specific to natural language processing, exBERT[3] can be used for visualizing self-attention patterns in Transformers, and AllenNLP[4] is useful for a variety of gradient-based methods. For generalized and model-agnostic suites of tools and techniques, we encourage interested readers to explore Lime[5], SHAP[6], IBM Explainability 360[7], ALIBI[8], Skater[9], iNNvestigate[10]

---

[1] captum.ai
[2] github.com/sicara/tf-explain
[3] github.com/bhoov/exbert
[4] allennlp.org/interpret

## III. DISCUSSION

The works included in this review span a wide range of sequential applications of deep learning involving a variety of sequential data types. While differing in execution, shared amongst all techniques is the desire to improve some aspect of human interpretability of the traditionally opaque deep models.

### A. Goal-Oriented Taxonomy

At current time, the notion of *interpretability* in machine learning remains vague and precisely undefined. In an effort to further the discussion and provide a more concrete foundation for integration with existing sequential applications, we broadly categorized methods according to the nature of the additional information which they make available. Given the rise of deep sequential learning, its recent trend towards human-critical applications, and the corresponding renewed research focus on explainability, we feel a goal-oriented classification of techniques is best suited for practitioners wishing to incorporate aspects of interpretability that are often domain and task specific.

A more traditional view of interpretability is embodied by the methods in our "Network Analysis" section, where the goal is to provide additional understanding of the mechanisms by which a deep sequential model transforms data between its two interpretable endpoints. Such techniques can be viewed as improving the transparency of the black box itself. Methods like weight inspection and maximum activation allow practitioners to understand the patterns and subsequences in the input space that each component of a deep neural network is trained to respond to. In essence, network analysis involves the notion of input/output correlations performed at a neuron, layer, or model-wise perspective and provide insight on how particular predictions are made on a more global scale. These techniques are not only useful for tracing the flow of information through a deep sequential network, but they also provide a clear view of the sequential pattern differences a model has learned between targets in the form of class prototypes.

---

[5] github.com/albermax/innvestigate
[6] github.com/slundberg/shap
[7] aix360.mybluemix.net
[8] github.com/SeldonIO/alibi
[9] github.com/oracle/Skater
[10] github.com/albermax/innvestigate

| Interpretability Goal | Interpretability Method | Requires Model Change? | Sequential Implementations |
|---|---|---|---|
| Network Analysis | Weight inspection | No | Karpathy et al. [18]<br>Lasko et al. [19]<br>Mehrabi et al. [20]<br>Li et al. [21]<br>Wang et al. [31] |
| | Maximum activation | No | Hermans & Schrauwen [23]<br>Karpathy et al. [24]<br>Dong et al. [25]<br>Che et al. [26]<br>Kale et al. [27]<br>Lanchantin et al. [28], [29]<br>Siddiqui et al. [30] |
| Sequential Saliency | Class activation maps | No | Wang et al. [31]<br>Lanchantin et al. [28,29] |
| | Gradient-based saliency | No | Lanchantin et al. [28,29]<br>Siddiqui et al. [30]<br>*Wavelet Decomposition Network*, Wang et al. [32] |
| | Attention | Yes | Chorowski et al. [67]<br>*Dipole,* Ma et al. [69]<br>*GRNN-HA,* Sha & Wang [70]<br>*Patient2Vec,* Zhang et al. [71]<br>*RETAIN,* Choi et al. [72], [73]<br>*RAIM,* Xu et al. [74]<br>*DA-RNN,* Qin et al. [75]<br>*DeepSOFA,* Shickel et al. [76]<br>*GeoMAN,* Liang et al. [77]<br>*HA-TCN,* Lin et al. [78] |
| | Perturbation and sensitivity | No | Alvarez-Melis & Jaakkola [45]<br>DeepBind, Alipanahi et al. [9] |
| | Sequential masking | Yes | *RCNN,* Lei et al. [83], [85]<br>Choi et al. 86]<br>Goh et al. [47] |
| | Erasure | No | Li et al. [46] |
| | Causality | No | Kale et al. [27] |
| | Sequential decomposition | No | Murdoch et al. [48] |
| | Timestep decay | Yes | Lei et al. [83]<br>*GRU-D,* Che et al. [84] |
| Multivariate attribution | Attention | Yes | *RETAIN*, Choi et al. [72], [73]<br>*DA-RNN*, Qin et al. [75]<br>*GeoMAN*, Liang et al. [77]<br>*DeepSOFA*, Shickel et al. [76]<br>*MV-LSTM*, Guo et al. [81] |
| | Backwards neuron attribution | No | *Layer-wise relevance propagation*, Bach et al. [33], [34], [35]<br>*DeepLIFT,* Shrikumar et al. [36], [37] |
| | Feature decomposition | Yes | Hsu et al. [79]<br>He et al. [84]<br>*MV-LSTM*, Guo et al. [81]<br>*IMV-LSTM*, Guo et al. [82] |
| | Feature decay | Yes | *GRU-D*, Che et al. [84] |
| | Model approximation | No | *Mimic learning*, Che et al. [38], [39], [40]<br>Wu et al. [41]<br>Murdoch & Szlam [42]<br>Krakovna & Doshi-Velez [43]<br>Schirrmeister et al. [44] |

*Table 1. Overview and implementation frameworks of goal-based taxonomy of sequential interpretability methods.*

In contrast to more dataset-wide insights from network analysis, the goal of techniques in our "Sequential Saliency" section focus on understanding relevant subsequences of a *single* sample that drive its output prediction. While network analysis can provide a view of input patterns that various regions of a deep learning model use to construct an overall sequence representation, they do not draw target-specific conclusions regarding important patterns between a single input and its corresponding output. In effect, many network analysis techniques can be viewed as an unsupervised form of interpretability using a supervised learning model. On the other hand, sequential saliency is solely focused on corelating exact patterns of an input to the prediction class. Such techniques often take the form of highlighting or visualizing subsequences of an input that are most responsible for its prediction. If a practitioner's goal is to provide an explanation for a prediction, sequential saliency methods are those that apply.

While the two aforementioned categories of interpretable methods are straightforward to understand in a univariate setting such as text or a single time series, many sequential applications involve multiple data points at each time step. While methods involving sequential saliency can discover which time steps were most influential, they are unable to identify and disentangle the importance of individual input variables at the salient time steps. In our section "Multivariate Attribution", we described techniques for isolating and quantifying importance at a per-dimension scale. These individual dimensions can refer to either the input variables themselves or such specific dimensions of a model's hidden internal representation.

## B. Comparison with Existing Interpretability Classifications
In the context of recently published overviews of interpretable machine learning methods[12,93] that tend to divide approaches as *intrinsic* or *post-hoc*, the majority of current techniques for sequential interpretability are decidedly *post-hoc* in nature; that is, given an already-trained deep learning model, these methods seek an understanding of what exactly a model has learned by either examining trained model parameters or by measuring the model's response to particular input sequences. This contrasts with Lipton's notion of *transparency*, which can generally be viewed as a complete understanding of the mechanistic data transformations occurring within the model itself. Exemplified by the relative lack of research in this area, precisely tracing inputs to outputs in deep learning frameworks is inherently difficult due to the fundamental nonlinear and hierarchical data transformations. While some may argue that post-hoc techniques cannot be fully trusted because they still rely on deep learning as a black box, we note Lipton's analogy[12] that if humans are considered interpretable, it is only in a post-hoc manner; while it is impossible to completely understand the precise biological mechanisms of human brain activity leading to an external expression or decision, it may not be strictly necessary, as useful information can still be gleaned from a variety of explanatory methods.

## C. Limitations and Future Direction
In this work we have provided an overview of current techniques that can be used to better understand deep learning frameworks in the context of sequential data types. After a comprehensive analysis of the body of related literature, we have identified several data-driven limitations and opportunities for future research that can be categorized as pertaining to either (1) data modalities, (2) application domains, and (3) interpretable techniques. These limitations represent a gap in current literature that are either natural extensions of existing techniques, or logical extensions of existing ideas and techniques.

Based on the contents of this review, we recommend further research and exploration pertaining to these issues.

Referring to Figure 2, the most common data type among deep sequential interpretability literature is text (29 studies), followed by time series (24), discrete sequences (13), and video (1). Given the prevalence of deep learning for video processing, it is surprising to see the lack of interpretable frameworks for this modality.

Additionally, while several works on this review aim to identify influential time steps of sequential input sequences, nearly all are viewed in isolation; that is, salient subsequences are identified by contiguous, single salient time steps rather than by directly modeling a sequence as a series of arbitrary length subsequences.

Missing from current literature are sequential applications of several domain and model-agnostic approaches to deep learning interpretability, such as local interpretable model-agnostic explanations (LIME)[94,95], meta learning[96], deep nearest neighbors[97], and adversarial techinques[98,99]. Such methods are general in nature, and while there exists no theoretical obstacle for sequential data types, at current time these techniques have not been explored or demonstrated for sequential data and applications.